\documentclass[sigconf, nonacm]{acmart}
\AtBeginDocument{%
  }

\usepackage{booktabs}
\usepackage{todonotes}
\usepackage{multirow}
\usepackage{algorithm}
\usepackage{algpseudocode}
\usepackage[most]{tcolorbox}
\usepackage{makecell}

\begin{document}

\title{Scalable Parameter-Light Spectral Method for Clustering Short Text Embeddings with a Cohesion-Based Evaluation Metric}


\author{Nikita Neveditsin}
\affiliation{%
  \institution{Saint Mary's University}
  \city{Halifax}
  \state{Nova Scotia}
  \country{Canada}
}
\email{nikita.neveditsin@smu.ca}

\author{Pawan Lingras}
\affiliation{%
  \institution{Saint Mary's University}
  \city{Halifax}
  \state{Nova Scotia}
  \country{Canada}
}
\email{pawan.lingras@smu.ca}

\author{Vijay Mago}
\affiliation{%
  \institution{York University}
  \city{Toronto}
  \state{Ontario}
  \country{Canada}
}
\email{vmago@yorku.ca}


\begin{abstract}
Clustering short text embeddings is a foundational task in natural language processing, yet remains challenging due to the need to specify the number of clusters in advance. We introduce a scalable spectral method that estimates the number of clusters directly from the structure of the Laplacian eigenspectrum, constructed using cosine similarities and guided by an adaptive sampling strategy. This sampling approach enables our estimator to efficiently scale to large datasets without sacrificing reliability. To support intrinsic evaluation of cluster quality without ground-truth labels, we propose the \emph{Cohesion Ratio}, a simple and interpretable evaluation metric that quantifies how much intra-cluster similarity exceeds the
global similarity background. It has an information-theoretic motivation
inspired by mutual information, and in our experiments it correlates
closely with extrinsic measures such as normalized mutual information and homogeneity. Extensive experiments on six short-text datasets and four modern embedding models show that standard algorithms like K-Means and HAC, when guided by our estimator, significantly outperform popular parameter-light methods such as HDBSCAN, OPTICS, and Leiden. These results demonstrate the practical value of our spectral estimator and Cohesion Ratio for unsupervised organization and evaluation of short text data. Implementation of our estimator of $k$ and Cohesion Ratio, along with code for reproducing the experiments, is available at \url{https://anonymous.4open.science/r/towards_clustering-0C2E}.

\end{abstract}




\maketitle





\section{Introduction}

Clustering short texts, represented as dense embeddings from transformer models~\cite{reimers2019sentence}, is a foundational task for organizing semantically related information in domains from social media analysis to digital health. A critical and persistent challenge in this domain, however, is the determination of the number of clusters, \emph{k}. Most clustering algorithms either require \emph{k} to be specified a priori or depend on other sensitive hyperparameters~\cite{campello2013density, frey2007clustering}, which is impractical in exploratory settings where ground-truth structure is unknown. This limitation severely hinders the application of clustering for unsupervised knowledge discovery~\cite{handl2005validation, park2019auto}.

To address the challenge of estimating the number of clusters without prior specification, we extend the classical eigengap heuristic from spectral graph theory~\cite{ng2002spectral, Luxburg2007SpectralClustering} to high-dimensional text embeddings. Our method constructs a normalized graph Laplacian from cosine similarities and applies adaptive thresholding to detect the eigengap, yielding a data-driven estimate of \emph{k} without manual tuning. A random subsampling strategy ensures scalability on large datasets, improving robustness across diverse text collections.

In addition, to evaluate clustering quality in the absence of ground-truth labels, we propose the \textbf{\emph{Cohesion Ratio}}, a simple and interpretable intrinsic metric. This metric evaluates a clustering's quality by comparing the average intra-cluster similarity against the global average similarity of the entire dataset. It is designed to reward semantically coherent clusters and, as we will show, correlates strongly with human-based extrinsic evaluations.

To validate our proposed methods, we conduct a large-scale empirical study on six diverse short-text datasets and four modern embedding models. We use our spectral method to guide traditional algorithms (K-Means and HAC) and compare their performance against prominent low-configuration methods, including HDBSCAN, OPTICS, Leiden, and Affinity Propagation. Our experiments are designed to answer two central questions: (i) can a robust spectral estimator enable traditional algorithms to outperform modern parameter-light methods? and (ii) how well do intrinsic metrics, including our proposed Cohesion Ratio, align with human-annotated labels?

This study makes the following key contributions:

\begin{itemize}
	\item \textbf{A Refined Spectral Method for Cluster Estimation:} Building on foundational work in spectral clustering, we introduce a refined and scalable method for estimating $k$. Our approach adapts the classic eigengap heuristic for high-dimensional text embeddings through an adaptive thresholding scheme and a sampling strategy inspired by recent large-scale estimation techniques~\cite{Fang2012BootstrapClusters, Estiri2018Kluster, Mahmud2023EnsembleClusters}. 

    \item \textbf{The Cohesion Ratio Metric:} We introduce a simple, interpretable, and effective intrinsic metric for evaluating cluster quality. Empirical results demonstrate that it correlates strongly with extrinsic, label-based metrics grounded in mutual information.

    \item \textbf{Extensive Empirical Validation:} Through rigorous experiments, we show that our spectral estimation method enables standard algorithms like HAC and K-Means to significantly outperform dedicated low-configuration methods on short-text clustering tasks. This provides direct, practical guidance for researchers and practitioners. 
\end{itemize}

\section{Related Work}  

\subsection{Exploratory Text Clustering When Number of Clusters is Unknown}
Clustering of text representations has been widely explored in natural language processing, particularly in the context of topic discovery, concept organization, and data profiling. Traditional approaches rely on models such as Latent Dirichlet Allocation \cite{blei2003latent} or methods operating on a vector space model of term frequencies, which infer latent topics by modeling word co-occurrence statistics. While effective on large corpora, these methods require substantial preprocessing, hyperparameter tuning (e.g., number of topics), and do not generalize well to short or sparse texts. Some attempts to address the unknown cluster count problem within this paradigm introduced their own limitations. For instance, the online clustering scheme proposed by Yin and Wang \cite{yin2016text} still required the user to specify a maximum possible number of clusters, \(k_{max}\).

In contrast, modern embedding-based clustering methods operate directly on dense vector representations produced by pretrained models. Sentence-level embedding models such as Sentence-BERT \cite{reimers2019sentence}, E5 \cite{wang2022text}, and more recent instruction-tuned transformers have been shown to yield high-quality representations across tasks. The Massive Text Embedding Benchmark (MTEB) \cite{muennighoff2023mteb} includes clustering tasks using K-Means with known cluster counts, providing insight into relative model performance. However, such evaluations assume prior knowledge of the number of clusters, which, as noted, remains a persistent challenge in real-world settings. This issue is not merely an inconvenience but a fundamental barrier to exploratory data mining, a challenge articulated compellingly by Keogh et al. \cite{keogh2004towards}, who argued for a community-wide shift towards parameter-free algorithms to enable more objective and reproducible data analysis.

\subsection{Parameter-Light Algorithms}
Density-based methods \cite{ester1996density} such as OPTICS \cite{Ankerst1999OPTICS} and HDBSCAN \cite{campello2013density} and message-passing approaches like Affinity Propagation \cite{frey2007clustering} attempt to remove the need for a user-specified $k$ by exposing alternative hyper-parameters such as minimum cluster size, preference, or reachability radius.  
Community-detection algorithms such as Louvain and Leiden transfer this idea to similarity graphs, using resolution parameters instead of $k$ and offering strong scalability on large corpora \cite{traag2019from}. However, each of the methods have their own shortcomings. Density-based methods are known to be sensitive to parameter selection \cite{dbscanlimit}, Affinity Propagation struggles on larger datasets due to its quadratic complexity, and community-detection algorithms such as Louvain and Leiden are known to sometimes produce poorly connected communities and encounter scalability limitations on massive graphs \cite{communlimit}.

\subsection{Spectral Clustering and Estimating the Number of Clusters}

Spectral clustering constructs a graph from the similarity matrix and derives its Laplacian, whose spectrum reveals underlying community structure \cite{ng2002spectral, Luxburg2007SpectralClustering}.  
A widely used method for estimating the number of clusters $k$ is the \emph{eigengap heuristic}, which selects $k$ based on the largest gap between consecutive eigenvalues.  
While conceptually simple and computationally efficient, this heuristic often breaks down when clusters differ significantly in scale or density.  

To address its limitations, several refinements have been proposed.  
Self-tuning spectral clustering replaces a global scale parameter with locally adaptive bandwidths \cite{zelnik2004self}.  
The Normalised Maximum Eigengap approach jointly optimizes graph parameters and $k$ \cite{park2019auto}, while the Spectral Information Criterion formalizes the ``elbow'' method using information-theoretic bounds \cite{martino2023spectral}.  
More recently, Yu et al.\ \cite{yu2025local} introduced fuzzy spectral clustering, which incorporates a fuzzy index to reduce sensitivity to the similarity matrix and simultaneously learns the clustering structure.

Despite these advances, estimating $k$ remains a persistent challenge, particularly in high-dimensional or semantically rich settings such as textual embeddings.  
Our work builds on this shortcoming by proposing a more robust and interpretable estimation technique tailored specifically to this domain.

\section{Methodology} 
\subsection{Problem Formulation}

Let $\mathcal{D} = \{x_1, x_2, \dots, x_n\}$ denote a collection of short text segments, and let $\Phi: \mathcal{D} \rightarrow \mathbb{R}^d$ be a pretrained embedding function mapping each $x_i$ to a dense vector $\mathbf{z}_i = \Phi(x_i)$. The objective is to partition the embedding set $\{\mathbf{z}_1, \dots, \mathbf{z}_n\}$ into semantically meaningful clusters using a function $f: \mathbb{R}^d \rightarrow \mathbb{N}$, without assuming prior knowledge of the number of clusters $k$ or relying on extensive hyper-parameter tuning. Here, \emph{semantically meaningful} refers to clusters whose members convey similar underlying topics, intents, or concepts, even if they differ lexically or syntactically.

This problem setting reflects real-world exploratory scenarios where human-aligned semantic groupings must be discovered without supervision. It presents two core challenges: (i) estimating the number of clusters $k$ in a scalable and robust way, and (ii) evaluating clustering quality intrinsically, without reference to external labels.

\subsection{Datasets}

We evaluate clustering methods on six diverse short-text datasets spanning a range of domains, including encyclopedic knowledge, news articles, conceptual categories, and user-generated discussions. Each dataset is standardized into a \texttt{text}-\texttt{label} format, where labels correspond to gold-standard semantic categories. To ensure fair evaluation across different sizes and label distributions, we apply both random and stratified sampling at fixed size intervals, resulting in a total of 62 dataset variants.

The DBpedia datasets consist of short titles and longer article bodies from the DBpedia 14 collection, each covering 14 uniformly distributed categories \cite{muennighoff2023mteb}. The 20 Newsgroups corpus contains over 11,000 long-form documents grouped into 20 topical categories with moderate imbalance \cite{Lang1995}. BLESS is a compact concept categorization benchmark with 200 noun entries distributed across 17 fine-grained semantic classes \cite{Baroni2011}. Finally, we include two large-scale user-generated datasets from MTEB: Reddit and StackExchange. These cover 50 and 121 categories respectively, and are characterized by moderate-length texts and substantial label imbalance \cite{muennighoff2023mteb}.

For each dataset, we compute summary statistics including the number of instances, number of classes, average input length (in words), and cluster imbalance (measured as the coefficient of variation of label counts). Table~\ref{tab:dataset_stats} presents the full statistics for the original datasets.

\begin{table}[h]
\centering
\caption{Full dataset statistics}
\resizebox{\linewidth}{!}{%
\begin{tabular}{lrrrrr}
\toprule
\textbf{Dataset} & \textbf{Instances} & \textbf{Clusters} & \textbf{Avg. Len. (Words)} & \textbf{Imbalance} \\
\midrule
DBpedia Title    & 560{,}000 & 14  & 2.74   & 0.000 \\
DBpedia Text     & 560{,}000 & 14  & 46.13  & 0.000 \\
20 Newsgroups    & 11{,}314  & 20  & 185.83 & 0.100 \\
BLESS            & 200       & 17  & 1.00   & 0.360 \\
Reddit           & 420{,}464 & 50  & 11.07  & 0.179 \\
StackExchange    & 373{,}850 & 121 & 9.67   & 0.400 \\
\bottomrule
\end{tabular}
}
\label{tab:dataset_stats}
\end{table}

\subsection{Embedding Models}

To represent short texts in a dense semantic space, we employ a selection of pretrained sentence embedding models with diverse architectures and training objectives. All models are openly available and can be used without fine-tuning. This diversity enables us to evaluate the robustness of clustering methods across embedding spaces that differ in scale, origin, and linguistic capabilities.

Our selected models include both compact encoders suitable for low-resource scenarios and larger instruction-tuned or retrieval-optimized architectures. Table~\ref{tab:embedding-models} summarizes the models along with their parameter sizes and reported performance on the Massive Text Embedding Benchmark (as of July, 2025).

\begin{table}[h]
\centering
\caption{Embedding models used for clustering. Parameter counts are given in millions (M) or billions (B). MTEB ranks are for English tasks as of July 2025}
\label{tab:embedding-models}
\resizebox{\linewidth}{!}{%
\begin{tabular}{@{}lcc@{}}
\toprule
\textbf{Model} & \textbf{Params} & \textbf{MTEB Rank} \\
\midrule
\makecell[l]{\texttt{multilingual-e5-large-instruct}}
& 560M & 25 \\
\makecell[l]{\texttt{Qwen3-Embedding-0.6B}}
& 600M & 4 \\
\makecell[l]{\texttt{Qwen3-Embedding-8B}}
& 8B & 2 \\
\makecell[l]{\texttt{Linq-Embed-Mistral}}
& 7B & 5 \\
\bottomrule
\end{tabular}%
}
\end{table}

The models were chosen to balance accessibility, representational diversity, and empirical competitiveness. This selection allows us to disentangle clustering algorithm behavior from embedding model biases and to test generalizability across small, multilingual, and instruction-optimized transformers.

\subsection{Clustering Algorithms}

To evaluate the effectiveness of our proposed cluster count estimator, we apply it in combination with two standard algorithms that require $k$ as input: \emph{K-Means} and \emph{Hierarchical Agglomerative Clustering (HAC)}. These algorithms are widely used and provide a controlled setting for isolating the impact of $k$ estimation on clustering performance.

For comparison, we include a set of representative \emph{low-configuration} algorithms that infer the number of clusters automatically or are inherently non-parametric. These serve as competitive baselines and fall into three categories: \emph{density-based}, \emph{graph-based}, and \emph{similarity-based} methods.

Among the density-based methods, HDBSCAN generalizes DBSCAN via hierarchical density estimation, constructing a condensed cluster tree and extracting a flat clustering with minimal parameter tuning. OPTICS similarly builds on the DBSCAN framework but uses reachability plots to handle variable-density structures without requiring a global threshold.

In the graph-based category, we employ Leiden clustering on similarity graphs constructed from cosine similarities between embeddings. The algorithm detects communities using the Constant Potts Model (CPM) \cite{cpm}, which promotes the discovery of well-connected subgraphs without needing to specify $k$.

Finally, Affinity Propagation adopts a message-passing strategy to identify exemplar points and induce clusters based on pairwise similarities. Like the other baselines, it does not require $k$ to be predefined.

This setup allows us to assess whether traditional algorithms, when guided by our spectral $k$-estimator, can outperform modern parameter-light methods under low-supervision constraints. Appendix~\ref{apx:algos} provides details on algorithm usage.

\subsection{Adaptive Estimation of the Number of Clusters via Laplacian Spectrum}

We propose a method to estimate the number of clusters in a dataset by analyzing the eigenspectrum of a normalized Laplacian derived from the cosine similarity of the samples. The approach combines spectral gap detection with adaptive sampling to remain efficient on large datasets.

Our method builds upon the well-established \emph{eigengap heuristic} from spectral graph theory, which suggests that a large difference between consecutive eigenvalues of the normalized Laplacian matrix corresponds to a natural cluster boundary in the data \cite{ng2002spectral, Luxburg2007SpectralClustering}. Formally, an ideal graph with \( k \) disconnected components has exactly \( k \) zero eigenvalues of the Laplacian \cite{Luxburg2007SpectralClustering}.

However, real-world similarity graphs derived from sentence embeddings are typically noisy, dense, and lack clearly disconnected components, which leads to a smoothed Laplacian spectrum and weakens the eigengap signal. To mitigate this, our estimator incorporates a moving-average normalization over the spectrum and applies adaptive sampling to increase robustness at scale.

\paragraph{Estimation Procedure}
Let \( \mathbf{X} \in \mathbb{R}^{n \times d} \) denote the matrix of embeddings for a dataset with \( n \) samples, where each sample has been mapped to a \( d \)-dimensional dense vector using the embedding function \( \Phi \) described earlier. We compute the cosine similarity matrix \( \mathbf{S} \in \mathbb{R}^{n \times n} \) and clip its entries to ensure non-negative affinities. Since negative cosine similarities are rare in semantic embedding spaces and usually lack meaningful interpretation \cite{cann2025using, ethayarajh-2019-contextual, jm3}, we set \( S_{ij} \leftarrow \max(S_{ij}, 0) \). We then construct the normalized Laplacian:
\[
\mathbf{L}_{\text{sym}} = \mathbf{I} - \mathbf{D}^{-1/2} \mathbf{S} \mathbf{D}^{-1/2},
\]
where \( \mathbf{D} \) is the diagonal degree matrix with \( D_{ii} = \sum_j S_{ij} \). The eigenvalues \( \lambda_1 \leq \lambda_2 \leq \dots \leq \lambda_n \) of \( \mathbf{L}_{\text{sym}} \) are then analyzed. A significant jump in the spectrum, often referred to as an \emph{eigengap}, is traditionally taken as an indicator of the number of clusters. However, in high-dimensional semantic spaces, the spectrum tends to decay smoothly, making it difficult to identify a single dominant gap.

Rather than seeking the largest eigengap, our method identifies the point at which the spectrum begins to \emph{flatten}, indicating that further eigenvalues contribute little additional structure. To detect this transition, we employ a moving average normalization strategy. For a given window size \( w \), we define the normalized spectral difference at position \( i \) as: 
\[
\delta_i = \frac{|\lambda_i - \lambda_{i-1}|}{\frac{1}{w} \sum_{j=i-w}^{i-1} \lambda_j + \epsilon},
\]
for \( i = w+1, \dots, n \), where \( \epsilon \) is a small constant to ensure numerical stability. We compute these normalized differences \( \delta_i \) across the upper half of the spectrum (i.e., for indices \( i > n/2 \), since we do not expect the number of clusters to exceed half the sample size), and denote their sample mean and standard deviation by \( \mathbb{E}[\delta] \) and \( \sigma[\delta] \), respectively.

Notably, preliminary experiments on our text embedding datasets reveal that standard eigengap heuristics consistently return values of \( k < 5 \), even when the ground truth number of categories is substantially larger. This behavior supports the interpretation that classic eigengap heuristics are ill-suited for spectral analysis over dense semantic graphs, where signal is distributed more gradually across the spectrum. Thus, we traverse the spectrum in reverse, searching for the first index \( i \) such that:
\[
\delta_i > \mathbb{E}[\delta] \left(1 + \frac{ \sigma[\delta]}{\mathbb{E}[\delta] + \epsilon} \right).
\]

\begin{figure}[h]
    \centering
    \includegraphics[width=1\linewidth]{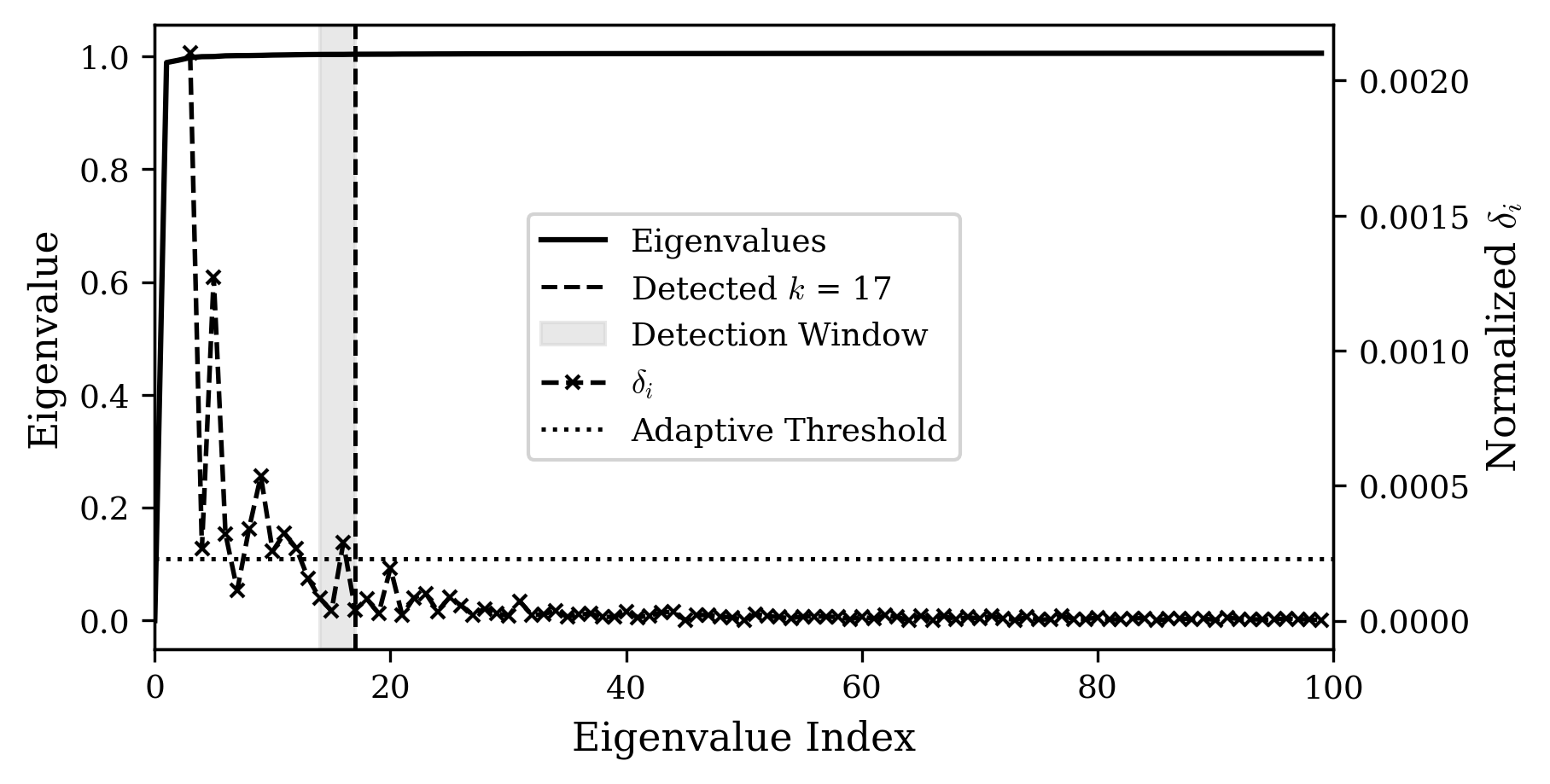}
    \caption{First half of Laplacian eigenvalue spectrum and normalized spectral differences (\( \delta_i \)) for the \textsc{BLESS} dataset. The detected number of clusters \( \hat{k} \) is shown as a vertical dashed line. The detection window (light gray) marks the region used for local averaging before the detected jump. The dotted line represents \( \delta_i \) values, and the horizontal dashed line indicates the adaptive threshold computed from the mean and variance of spectral differences. A significant over-threshold jump at \( \hat{k} \) signals the point where spectral flattening begins, suggesting the  cut-off for meaningful clustering.}
    
    \label{fig:demo}
\end{figure}

The multiplicative term reflects a data-dependent \textit{threshold} that scales with the relative variation in spectral differences. This formulation ensures that the selected point in the spectrum is not only larger than the average change but also exceeds typical fluctuations observed across the spectrum, capturing only meaningful transitions in spectral decay. An example of this detection process is shown in Figure~\ref{fig:demo}, which visualizes the Laplacian spectrum, normalized differences, and adaptive threshold for the \textsc{BLESS} dataset.

To scale this method to large datasets, we apply adaptive sampling: if \( n > \tau \), where \( \tau \) is a maximum sample size (e.g., 1000), we draw multiple random subsets and average the estimated number of clusters across replicates. We set the number of replicates as \( r = \log_2(n) \cdot 10 \), allowing the number of subsampled evaluations to scale smoothly with dataset size. This adaptive growth balances computational cost with the stability of cluster number estimation, similar to practices in bootstrap-based and ensemble clustering methods.
This ensemble approach is motivated by recent work in scalable cluster estimation, such as the \texttt{kluster} procedure \cite{Estiri2018Kluster}, bootstrap-based estimators \cite{Fang2012BootstrapClusters}, and ensemble eigengap estimation methods for big data \cite{Mahmud2023EnsembleClusters}. Algorithm~\ref{alg:adaptive_cluster_estimation} summarizes the procedure. 

\begin{algorithm}[h]
\caption{Adaptive Cluster Count Estimation}
\label{alg:adaptive_cluster_estimation}
\begin{algorithmic}[1]
\Require Data matrix \( \mathbf{X} \in \mathbb{R}^{n \times d} \), sample cap \( \tau \), window size \( w \), fallback default \( k_{\text{default}} = 5 \)
\If{ \( n \leq \tau \) }
    \State \( \Lambda \gets \texttt{ComputeEigenvalues}(\mathbf{X}) \)
    \State \Return \texttt{EstimateKFromSpectrum}(\( \Lambda, w, k_{\text{default}} \))
\Else
    \State \( r \gets \log_2(n) \cdot 10\)

    \For{ \( i = 1 \) to \( r \) }
        \State Draw random subset \( \mathbf{X}_i \subset \mathbf{X}, |\mathbf{X}_i| = \tau \)
        \State \( \Lambda_i \gets \texttt{ComputeEigenvalues}(\mathbf{X}_i ) \)
        \State \( k_i \gets \texttt{EstimateKFromSpectrum}(\Lambda_i, w, k_{\text{default}}) \)
    \EndFor
    \State \Return \( \frac{1}{r} \sum_{i=1}^r k_i \)
\EndIf
\end{algorithmic}
\end{algorithm}

The ComputeEigenvalues subroutine computes the cosine similarity matrix $\mathbf{S}$ from $\mathbf{X}$, rectifies to non-negative, forms symmetric normalized Laplacian, and returns sorted eigenvalues $\Lambda$. The EstimateKFromSpectrum subroutine computes normalized differences $\delta_i$ with window $w$ on eigenvalues $\Lambda$, sets the adaptive threshold, traverses reverse to find first $\delta_i >$ threshold for $\hat{k}$, defaulting to $k_{\text{default}}$  (chosen as a conservative prior for short‑text corpora) if none found.

\paragraph{Sensitivity Analysis for $w$ and $\tau$}
We evaluate the effect of the moving average window size $w$ on the accuracy of cluster count estimation. For each value of $w$, we compute the mean relative error between the estimated and true number of clusters across datasets, along with the average fraction of failed predictions returned by the estimator (when it needed to fall back to $k_{\text{default}}$).

\begin{figure}[h!]
  \centering
  \includegraphics[width=1\linewidth]{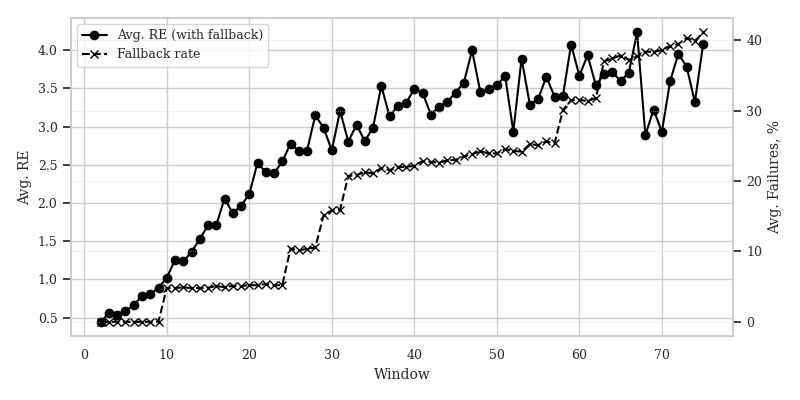}
  \caption{Impact of window size $w$ on mean relative error (solid line, left axis) and average fraction of invalid predictions (dashed line, right axis).}
  \label{fig:window-sensitivity}
\end{figure}

Figure~\ref{fig:window-sensitivity} summarizes these results. As the window size increases, the average relative error tends to grow, indicating reduced estimation precision due to oversmoothing. Notably, the fraction of invalid predictions also rises with larger windows, reflecting a loss in estimator selectivity. Moderate window sizes $w \lesssim 7$ offer the best trade-off between accuracy and robustness. These results support the use of $w=3$ as a default setting. 

Further, to balance estimation accuracy and computational cost, we conducted an ablation study across multiple values of the sampling threshold~$\tau \in \{200, 500, 1000, 2000\}$. Results showed that while larger values of~$\tau$ increase the runtime due to the~$\mathcal{O}(\tau^3)$ complexity of eigendecomposition, they do not consistently improve cluster estimation accuracy. Most datasets perform well in the~$500$--$1000$ range, with~$\tau=1000$ offering a reliable trade-off. Notably, smaller values were inadequate for datasets with many clusters, where the truncated spectrum lacked sufficient resolution to detect the spectral flattening point.

\paragraph{Optional Normalization Variant}
While our core algorithm operates directly on raw cosine similarities, we also performed an ablation study in which Z-score normalization (followed by rectification) was applied prior to Laplacian construction. This step aims to suppress weak inter-cluster affinities and highlight confident intra-cluster connections.

We evaluated the impact of Z-score normalization across a range of moving average window sizes \( w \in \{3, 4, 5, 6, 7\} \) and subsample caps \( \tau \in \{200, 500, 1000, 2000\} \). Overall, the Z-score variant yielded a lower mean relative error (0.6017) compared to the raw similarity baseline (0.7201), with a Wilcoxon signed-rank test indicating the difference was statistically significant (\( p < 0.05 \)). At the dataset level, Z-scoring led to significant improvements on four out of six datasets, while degrading performance on Reddit and StackExchange. 
Given these mixed outcomes, we do not include normalization in the default pipeline, but offer it as an optional preprocessing step that may benefit datasets with balanced and semantically well-separated clusters. Appendix~\ref{apx:zscoreraw} shows heatmaps of raw and normalized cosine similarity matrices for selected datasets, illustrating that normalization can reveal sharper intra-cluster blocks and reduce background noise in some cases.

\paragraph{Computational Complexity.}
We cap each subsample at size \( \tau \), and draw \( r = \log_2(n) \cdot 10 \) subsets, yielding total complexity \( O(r \cdot \tau^3) = O(\log n \cdot \tau^3) \), dominated by the eigendecomposition of \( \tau \times \tau \) similarity matrices. This is substantially more efficient than full spectral clustering, which incurs \( O(n^3) \) time.

Our estimator is modular and algorithm-agnostic, making it suitable as a general-purpose \( k \)-estimation module for any clustering algorithm that benefits from a prior on the number of clusters.

Philosophically, our use of subsampling parallels fast approximate spectral clustering methods~\cite{yan2009fast}, though we apply it to the problem of estimating \( k \), not clustering itself. The strong performance of our method, particularly its stable relative error in cluster estimate across dataset sizes (see Section 4.3 for details) supports its viability as a scalable and reliable \( k \)-estimator.

\subsection{Extrinsic and Intrinsic Evaluation}

We evaluate the effectiveness of parameter-light clustering algorithms on sentence-level embeddings of short texts, comparing their performance against our proposed $k$-estimation method combined with K-Means and HAC. 

\subsubsection{Extrinsic Evaluation Metrics}

We evaluate clustering performance using three standard extrinsic metrics that quantify alignment with gold-standard labels:

\textbf{Normalized Mutual Information (NMI)}~\cite{strehl2002cluster,vmeasure}: Measures mutual dependence between predicted and true clusters, normalized to the range $[0,1]$. It is equivalent to the V-measure under symmetric normalization, which decomposes into \textit{homogeneity} (each cluster contains only members of a single class) and \textit{completeness} (all members of a class are assigned to the same cluster).

\textbf{Adjusted Rand Index (ARI)}~\cite{hubert1985comparing}: Quantifies pairwise agreement between clusterings, adjusted for chance. A value of 0 indicates random agreement; 1 indicates perfect alignment.

\textbf{Fowlkes-Mallows Index (FMI)}~\cite{fowlkes1983method}: Measures the geometric mean of precision and recall between pairs of points in predicted and true clusters. Values range from 0 (no agreement) to 1 (perfect match), with higher values indicating greater consistency between the predicted and ground-truth clustering structures.

\textbf{\( k \)-Relative Error}: Measures the deviation between the estimated number of clusters \( \hat{k} \) and the ground-truth count \( k \):
\[
\text{RE}_{k} = \frac{|\hat{k} - k|}{k}
\]

\subsubsection{Intrinsic Evaluation Metrics}

We assess clustering structure using intrinsic metrics that do not rely on ground-truth labels:


\textbf{Silhouette Score}~\cite{rousseeuw1987silhouettes}: Captures both intra-cluster cohesion and inter-cluster separation. Ranges from $-1$ to $1$, with higher values indicating more compact and well-separated clusters.

\textbf{Davies-Bouldin Index (DBI)}~\cite{Davies1979}: Computes average similarity between each cluster and its most similar peer. Lower values indicate better-defined clusters.

\textbf{Calinski-Harabasz Index (CHI)}~\cite{Calinski1974}: Measures the ratio of between-cluster to within-cluster dispersion. Higher values reflect more distinct and well-separated clusters.

\textbf{Cohesion Ratio.}
We propose a cohesion-based internal metric that evaluates the average intra-cluster similarity of a clustering solution relative to the overall similarity structure of the dataset. The core intuition is that a meaningful clustering should exhibit stronger internal coherence than the background similarity present across the entire dataset. Mathematically, this formulation adapts the Pairwise Discriminative
Power (PDP) statistic, originally introduced for supervised product-resolution
similarity evaluation \cite{balog2011investigation}. However, our application differs substantially in scope and intent: whereas PDP is defined with respect to ground-truth equivalence classes, we employ this ratio purely on the predicted clustering as a label-free intrinsic index for assessing the quality of clusterings over short-text embeddings. Let \( X = \{x_1, \dots, x_n\} \) be the set of data points, partitioned into clusters \( \mathcal{C} = \{C_1, \dots, C_K\} \), where each \( C_k \subset X \) is a subset of indices corresponding to a single cluster. Let \( S_{ij} \) denote a general similarity measure between points \( x_i \) and \( x_j \); in our implementation, we use cosine similarity.

We first compute the global average pairwise similarity across all unordered pairs of points:
\[
\mu_G = \frac{1}{\binom{n}{2}} \sum_{1 \le i < j \le n} S_{ij}.
\]
Next, we compute the average intra-cluster similarity, denoted \( \mu_I \), by summing the similarities between all unordered point pairs within each cluster and dividing by the total number of such pairs:
\[
\mu_I = \frac{1}{P} \sum_{k = 1}^K \sum_{\substack{i, j \in C_k \\ i < j}} S_{ij},
\]
where \( P = \sum_{k=1}^K \binom{|C_k|}{2} + N_1 \), and \( N_1 \) is the number of singleton clusters. Singleton clusters are treated as contributing a default similarity equal to the global average \( \mu_G \), and each contributes a single virtual pair to the total. This treatment ensures that singletons are neutral in the cohesion calculation, as they neither inflate nor penalize the score, reflecting their ambiguous contribution to the overall cluster structure.

The final cohesion ratio is defined as:
\[
\rho_C = \frac{\mu_I}{\mu_G}.
\]
Since \( S_{ij} \in [0, 1]\), both \(\mu_I\) and \(\mu_G\) lie in the interval \([0, 1]\). Therefore, the Cohesion Ratio \(\rho_C = \mu_I / \mu_G\) is well-defined and satisfies \(\rho_C \in [0, \infty)\), with \(\rho_C = 1\) corresponding to a clustering whose internal cohesion is equal to the global background similarity. In practice, most well-formed clusterings yield \(\rho_C > 1\), while poorly formed or random clusterings tend to have \(\rho_C \approx 1\). This boundedness makes \(\rho_C\) robust and interpretable across datasets with varying structure and density.

The ratio thus quantifies the relative strength of intra-cluster similarity compared to the background similarity structure of the data. It is agnostic to geometric shape, convexity, or distributional assumptions, making it broadly applicable across a range of domains and similarity measures. 

\paragraph{Theoretical Motivation via an Information-Theoretic Analogy}
We interpret $\rho_C$ as a proxy for the information gained by conditioning pairwise affinities on the predicted cluster assignments $C$. From a null-model perspective~\cite{hubert1985comparing}, the global average affinity $\mu_G$ represents the expected similarity under random assignments (background ``noise''), while $\mu_I$ reflects the concentration of affinity within clusters (signal).

To formalize this contrast, we model the affinity space as a binary channel with two states (``similar'' vs.\ ``not similar''). Let $A \in \{0,1\}$ denote a Bernoulli random variable indicating whether a randomly sampled pair is similar, induced from the underlying affinities $S_{ij}$. In this abstraction, we treat the global mean $\mu_G$ as the prior probability of similarity, $P(A{=}1) \approx \mu_G$, and the intra-cluster mean $\mu_I$ as the conditional probability given cluster agreement, $P(A{=}1 \mid C_{\mathrm{same}}) \approx \mu_I$.

The divergence between the cluster-conditioned model and the background null model can be expressed using the Kullback-Leibler divergence between two Bernoulli distributions:
\[
D_{\mathrm{KL}}(\mu_I \parallel \mu_G)
= \mu_I \log\left(\frac{\mu_I}{\mu_G}\right)
  + (1 - \mu_I) \log\left(\frac{1 - \mu_I}{1 - \mu_G}\right).
\]
In the regime typical for short-text clustering, where semantic connections are sparse ($\mu_G \to 0$) but clusters are coherent ($\mu_I \gg \mu_G$), the first term tends to dominate the divergence. This term corresponds to the expected log-likelihood ratio of observing a ``similar'' pair under the cluster hypothesis versus the null hypothesis.

Consequently, the logarithm of our Cohesion Ratio corresponds directly to the \textit{Pointwise Mutual Information} of this binary event:
\[
\log \rho_C = \log\left(\frac{\mu_I}{\mu_G}\right)
\approx \log \frac{P(A{=}1 \mid C_{\mathrm{same}})}{P(A{=}1)}
= \operatorname{PMI}(A{=}1, C_{\mathrm{same}}).
\]
Under this analogy, $\rho_C \approx 1$ implies $\operatorname{PMI} \approx 0$, meaning that conditioning on cluster membership provides no information gain regarding similarity. In contrast, $\rho_C \gg 1$ indicates substantial deviation from the null, corresponding to meaningful structural information. Empirically, our results confirm that $\rho_C$ correlates strongly with extrinsic information-theoretic measures such as normalized mutual information.

Appendix~\ref{apx:chvssilh} discusses differences between Cohesion Ratio and Silhouette Score.


\section{Results}
\subsection{Extrinsic and Intrinsic Evaluation}

Table~\ref{tab:overall_scores} compares the clustering algorithms across extrinsic and intrinsic evaluation metrics. The Friedman test with $k = 6$ clustering methods and $N = 248$ dataset-model pairs revealed statistically significant differences ($p \ll 0.05$) among the algorithms across all evaluation metrics.

\begin{table}[ht!]
\centering
\caption{Overall Scores for the Algorithms: Extrinsic and Intrinsic Evaluation. \textbf{Bold} values indicate the best score per column, while \textit{italic} values indicate the worst. The term \textit{k-est} refers to our method for estimating the number of clusters ($k$). }
\label{tab:overall_scores}
\resizebox{\linewidth}{!}{
\begin{tabular}{lrrrrrrrr}
\toprule
\textbf{Algorithm} & \textbf{ARI} & \textbf{NMI} & \textbf{F-M} & \textbf{$RE_k$} & \textbf{Silh.} & \textbf{DBI} & \textbf{CHI} & \textbf{$\rho$} \\
\midrule
HDBSCAN & 0.0163 & \textit{0.2661} & 0.1582 & 16.3348 & -0.0725 & 2.9791 & 3.8561 & 1.0057 \\
OPTICS & \textit{0.0045} & 0.3486 & \textit{0.1384} & \textit{33.3208} & \textit{-0.1216} & \textbf{1.7632} & \textit{2.1513} & \textit{0.9949} \\
Aff. Prop. & 0.1293 & 0.5730 & 0.1900 & 20.6124 & 0.0567 & 2.8967 & 5.9439 & \textbf{1.5112} \\
Leiden & 0.2767 & \textbf{0.6162} & 0.3278 & 9.0217 & 0.0264 & 2.0598 & 7.0905 & 1.3303 \\
K-Means (k-est) & 0.3534 & 0.5929 & 0.3916 & 0.5203 & \textbf{0.0664} & \textit{4.1446} & \textbf{53.1836} & 1.2969 \\
HAC (k-est) & \textbf{0.3799} & 0.5932 & \textbf{0.4175} & \textbf{0.5223} & 0.0597 & 4.1397 & 47.8207 & 1.2786 \\
\bottomrule
\end{tabular}
}
\end{table}

To assess the robustness of clustering performance differences, we conducted the Nemenyi post-hoc test at a significance level of $\alpha = 0.05$. The results indicate that both K-Means and HAC show statistically significant improvements over several other clustering algorithms across multiple evaluation metrics. Specifically, for the F-M score, HAC significantly outperforms all other methods. For the ARI score and $RE_k$, HAC significantly outperforms all other methods except K-Means, with which it shows no significant difference. 

For NMI, Leiden significantly outperforms all other methods. K-Means and HAC are statistically indistinguishable from each other and from Affinity Propagation. A deeper analysis of these differences shows that Affinity Propagation significantly outperforms all other methods in terms of homogeneity, with Leiden ranking second, and K-Means and HAC tied for third. The completeness score shows a different pattern, with K-Means, HAC, and Leiden ranked first, followed by Affinity Propagation. Notably, the Cohesion Ratio aligns with the statistical ranking of the Homogeneity score, while exhibiting moderate concordance with the Completeness score. In the latter case, HAC emerges as the clear leader, whereas Affinity Propagation outperforms only OPTICS and HDBSCAN.

For the Silhouette Score, K-Means and HAC are statistically indistinguishable from one another and both significantly outperform all other methods, except in the case where HAC does not differ significantly from Affinity Propagation. For the Calinski-Harabasz Index, K-Means significantly outperforms all other methods, with HAC ranked second.

These findings highlight the consistent and robust performance of the spectral estimate-based K-Means and HAC methods across both extrinsic and intrinsic clustering quality measures. Interestingly, Affinity Propagation demonstrates strong performance in terms of Homogeneity, Cohesion Ratio, and Silhouette Score, but underperforms on Completeness and exhibits a tendency to severely overestimate the number of clusters, which is reflected in its moderate extrinsic evaluation scores.

\subsection{Correlation Between Extrinsic and Intrinsic Evaluation Metrics}

\begin{figure}[htbp]
    \centering
    \includegraphics[width=1\linewidth]{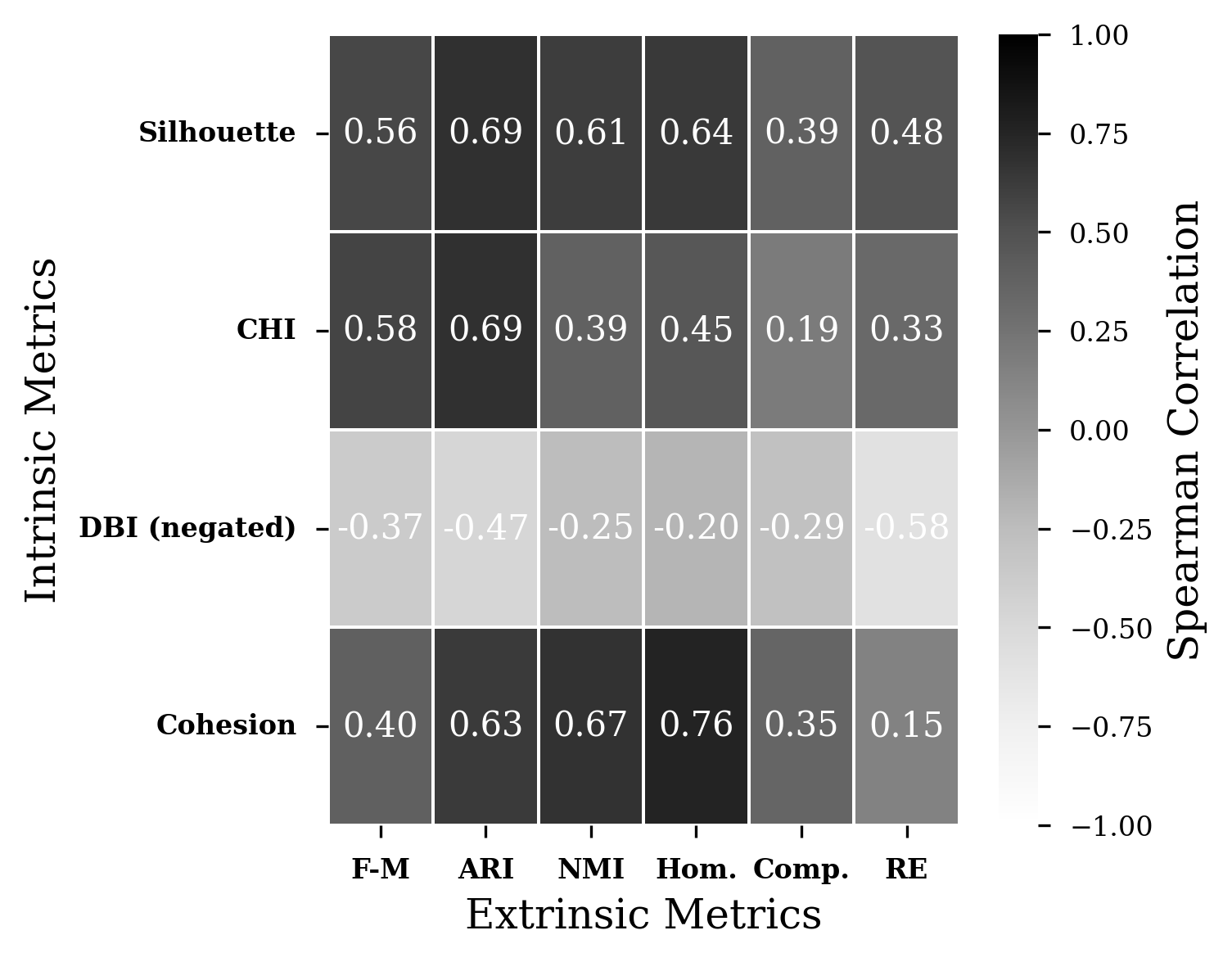}
    \caption{
        Spearman correlations between intrinsic and extrinsic clustering metrics. Higher values indicate stronger agreement 
        with extrinsic clustering scores.
    }
    \label{fig:correlation_heatmap}
\end{figure}

To further understand the relationship between internal cluster validity indices and external evaluation measures, we compute Spearman correlation coefficients across all clustering configurations. The resulting heatmap in Figure~\ref{fig:correlation_heatmap} illustrates the degree to which each intrinsic metric aligns with extrinsic clustering quality scores.

The correlation heatmap indicates that, while the Silhouette Score remains a widely used and effective intrinsic metric, exhibiting strong positive correlations with most extrinsic evaluation metrics, the proposed Cohesion Ratio demonstrates even greater alignment in several key areas when applied to clustering of text representations. In particular, Cohesion Ratio attains the highest correlation among all intrinsic metrics with Normalized Mutual Information and Homogeneity, and its correlation with Adjusted Rand Index is comparable to that of the Silhouette Score. These findings suggest that, in the context of clustering high-dimensional textual embeddings, the Cohesion Ratio is not merely a simpler alternative, but a robust and effective intrinsic metric that exhibits strong alignment with extrinsic measures grounded in mutual information. Interestingly, the Davies-Bouldin Index (negated, as lower values indicate better clustering) exhibits negative correlations with most extrinsic metrics. This suggests that DBI may be ill-suited for evaluating the clustering quality of dense text embeddings, likely due to its underlying assumption of convex and well-separated cluster structures, a condition that rarely holds in high-dimensional text embeddings.

\subsection{Relationship Between Predicted Clusters and True Classes, Scalability of Proposed Estimator}

\begin{figure}[h!]
    \centering
    \includegraphics[width=1\linewidth]{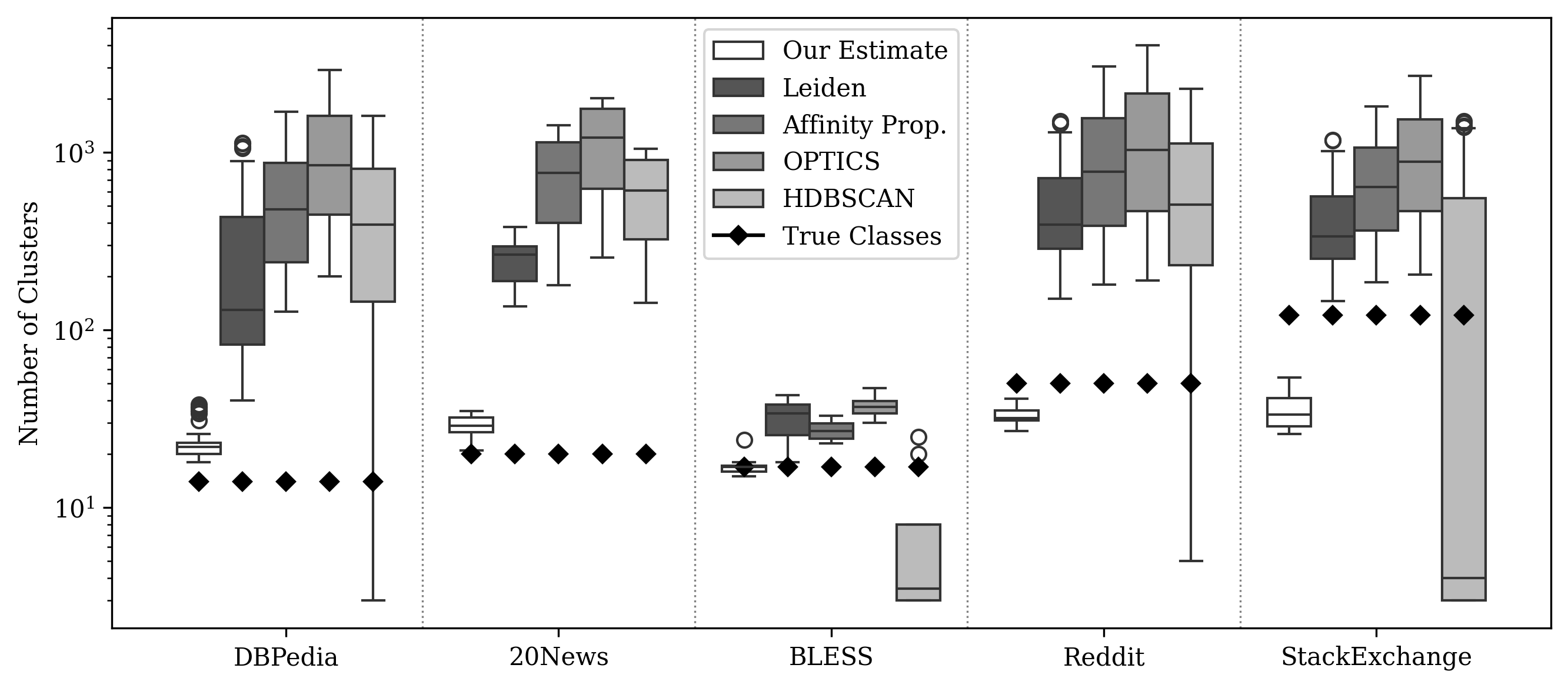}
    \caption{
        Predicted number of clusters (log scale) across stratified datasets for different clustering algorithms. Each box shows the distribution over multiple runs. Black diamonds indicate the true number of underlying classes. Our method tends to produce cluster counts closest to ground truth across most datasets.
    }
    \label{fig:estimates}
\end{figure}

Figure~\ref{fig:estimates} illustrates the distribution of predicted cluster counts across stratified datasets for a set of representative clustering algorithms.

Additional visualizations illustrating how evaluation metrics vary with the number of clusters are provided in Appendix~\ref{apx:profiles}. These plots offer insight into the relatively limited performance of the estimator on the StackExchange dataset. In particular, the curves for extrinsic evaluation metrics: ARI and F-M, are comparatively flat and low, indicating that current embedding models may inadequately capture the underlying structure of this dataset. Nonetheless, it is noteworthy that the estimated number of clusters closely corresponds to the peak of the Silhouette score, suggesting that the estimator remains sensitive to intrinsic structure. Furthermore, the visual alignment between the Cohesion Ratio and Homogeneity curves provides empirical support for the conceptual relationship between the Cohesion Ratio and mutual information.

To evaluate robustness to dataset size, we conducted a Kruskal-Wallis H-test to assess whether the estimated number of clusters using our method varies significantly across sizes within each dataset group. The null hypothesis assumes that the estimates come from the same distribution regardless of dataset size. Our evaluation spans four dataset groups: DBpedia Title, DBpedia Text, Reddit, and StackExchange. For each group, we tested our estimator on stratified subsets ranging from fewer than 1,000 instances to over 300,000 instances. The test returned a p-value of 0.429 for all groups, indicating no statistically significant difference. This supports the claim that our proposed estimator is both scalable and robust to variations in dataset size.


\section{Discussion and Conclusion}



In this study, we addressed the challenge of clustering short-text embeddings in parameter-light settings by proposing and evaluating a  scalable method for estimating the number of clusters ($k$). Our analysis demonstrates that this spectral estimation technique is highly effective. When paired with traditional algorithms like Hierarchical Agglomerative Clustering  and K-Means, it significantly outperforms established parameter-light methods across multiple datasets and evaluation criteria.  

Another key contribution of this work is the development and analysis of the
\emph{Cohesion Ratio} ($\rho_C$), a simple and interpretable label-free
intrinsic metric with an information-theoretic interpretation inspired by
mutual information.  Our findings validate its utility, showing that it has a stronger correlation with extrinsic metrics like Normalized Mutual Information and Homogeneity than traditional indices. These results position the Cohesion Ratio as a robust alternative to metrics like the Silhouette Score for assessing clusterings of high-dimensional text embeddings where ground-truth labels are absent.

Ultimately, this study provides practitioners with a validated, data-driven approach for clustering short texts without prior knowledge of $k$. The proposed spectral estimation method and the Cohesion Ratio offer powerful tools for the unsupervised organization and evaluation of textual data, for practitioners faced with clustering short-text data without prior knowledge of the number of clusters, our findings suggest that using our spectral estimator with a standard algorithm like HAC or K-Means is a more effective and robust strategy than relying on popular out-of-the-box parameter-light methods, paving the way for more robust and interpretable knowledge discovery.

\section{Limitations}

First, while we evaluate a diverse range of classical and parameter-light clustering methods, we do not include recent deep clustering \cite{xie2016dec} approaches in our benchmarks. These methods often require task-specific tuning for pretraining or fine-tuning, which contradicts our core objective: to develop practical, label-free tools for exploratory clustering of short text. 

Second, the ground-truth labels used for extrinsic evaluation may themselves reflect subjective or task-specific categorizations, introducing further interpretive variability.

Third, while our estimator performs well across several datasets, its robustness may depend on factors such as the number of clusters, cluster granularity, and semantic density. Our current evaluation is limited to a small set of high-quality short-text datasets, many of which fall within a narrow cluster range. Broader validation across more diverse and well-annotated datasets is needed, but further research is hindered by the limited availability of such resources.

\section{Further Research}

Future work can advance this study along three complementary lines. First, clustering objectives could be designed to directly optimize Cohesion Ratio, potentially balanced by auxiliary metrics such as the Silhouette Score to avoid over-fragmentation. Such objectives could be explored through heuristic search methods like simulated annealing~\cite{selim1991anneal} or via differentiable relaxations, opening the door to more intuitive and human-aligned cluster structures. Second, Cohesion Ratio could be embedded into recent unified clustering frameworks, such as the density-connectivity distance formulation of Beer et al.~\cite{Beer}, enabling parameter-light algorithms that bridge density-, centroid-, and spectral-based approaches within a principled formalism. Finally, the proposed spectral \(k\)-estimator can be developed further in several directions: devising adaptive mechanisms for selecting smoothing parameters \((w,\tau)\) based on stability or information criteria; quantifying estimator uncertainty with confidence intervals through subsampling or bootstrapping; improving robustness to different graph constructions (e.g., sparsification, normalization, or rectification choices). Together, these avenues promise both theoretical foundations and practical algorithms for more reliable, interpretable, and parameter-light clustering in high-dimensional embedding spaces.


\bibliographystyle{ACM-Reference-Format}


\appendix

\onecolumn
\newpage

\section{Configurations of algorithms}
\label{apx:algos}

\footnotesize
\begin{table*}[ht]
\centering
\caption{Configuration Details of Clustering Algorithms Used in the Study}
\begin{tabular}{|p{3.2cm}|p{3.5cm}|p{8.0cm}|}

\hline
\textbf{Algorithm} & \textbf{Configurable Parameters} & \textbf{Notes} \\
\hline
Affinity Propagation & preference (implicit default: mean similarity) & Uses default preference. \\
\hline
HDBSCAN & min\_cluster\_size=2 & Preliminary experiments with min\_cluster\_size values of 2, 5, and 10 showed that a value of 2 provided a slightly better balance between ARI and NMI across datasets, so we report results with this setting. \\
\hline
OPTICS & min\_samples=2, cluster\_method='xi' & Preliminary experiments with min\_samples=5 and 10 led to a noticeable degradation in NMI compared to 2, so we adopt the latter as the main configuration. \\
\hline
Leiden (CPM on Similarity Graph) & resolution\_parameter=1 (default setting) & We apply the Leiden algorithm for community detection on the cosine similarity graph with the resolution parameter fixed at its default value of \(1.0\).  
For constructing the graph, we tested two normalization strategies on the cosine similarities:  
(i) simple rectification \(S_{ij} \leftarrow \max(S_{ij},0)\), and  
(ii) z-score normalization followed by rectification.  
Preliminary experiments showed that approach (ii) produced slightly better results in terms of ARI and NMI across datasets, so we adopted it for the main experiments.
 \\
\hline
\end{tabular}
\label{tab:clustering_configs}
\end{table*}

\section{Similarity Matrices: Raw vs. Z-score Normalized and Rectified}
\label{apx:zscoreraw}

\begin{figure*}[h]
    \centering
    \includegraphics[width=1\linewidth]{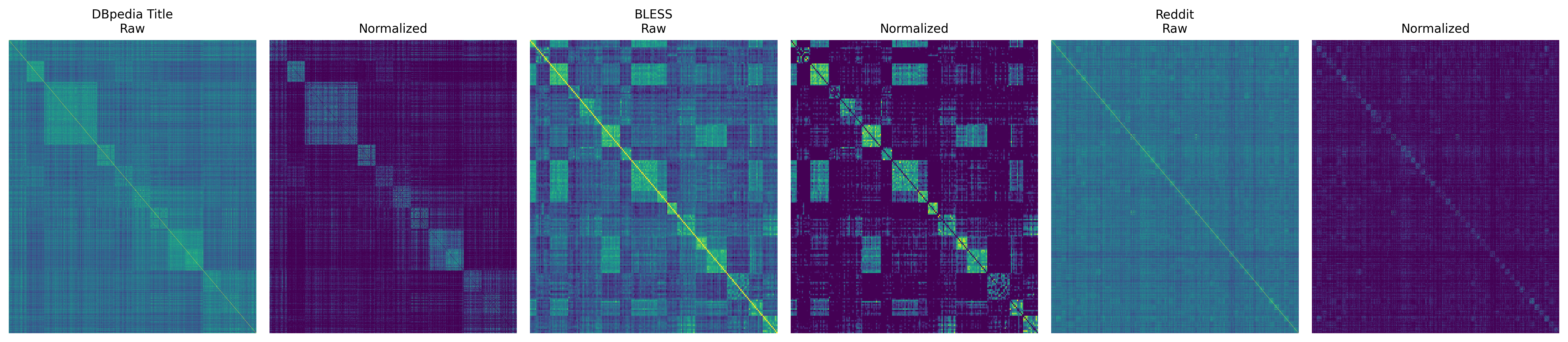}
    \vspace{1mm} 
    \includegraphics[width=1\linewidth]{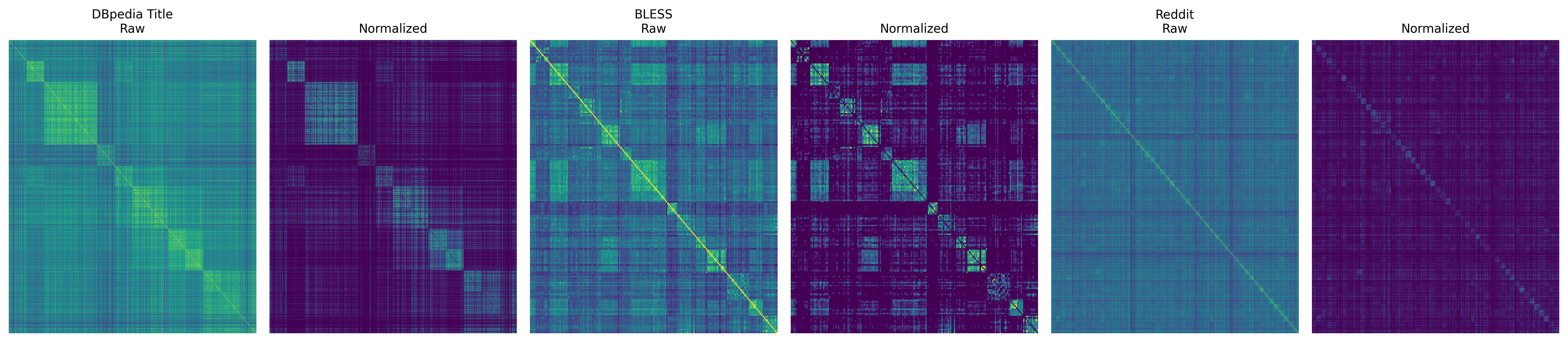}
\caption{
Heatmaps of cosine similarity matrices for BLESS dataset and two stratified samples of DBpedia and Reddit datasets. 
Each pair shows the (left) raw cosine similarities and (right) z-score normalized and rectified versions. 
The first row corresponds to embeddings from the \texttt{multilingual-e5-large-instruct} model (smallest model from our experiments), while the second row uses the \texttt{Qwen3-Embedding-8B} model (largest model from our experiments). 
Normalization emphasizes strong intra-cluster similarities while suppressing noise, resulting in clearer block-like structures aligned with true class boundaries in some cases.
}
    \label{fig:zscore-heatmaps}
\end{figure*}

\newpage
\normalsize
\section{Discussion on Cohesion Ratio vs Silhouette Score}
\label{apx:chvssilh}
While the Cohesion Ratio and the Silhouette Score both aim to quantify intra-cluster cohesion, they rely on fundamentally different formulations. In the Silhouette Score, cohesion is captured by the pointwise term \( a(i) \), defined as the average distance from a point \( x_i \) to all other points within its assigned cluster:
\[
a(i) = \frac{1}{|C_k| - 1} \sum_{\substack{x_j \in C_k \\ j \ne i}} \text{dist}(x_i, x_j),
\]
where \( C_k \) denotes the cluster containing \( x_i \), and \( \text{dist} \) is typically Euclidean distance. This formulation emphasizes the \emph{local geometric compactness} of each point's neighborhood and is sensitive to the shape, size, and boundary structure of individual clusters.

In contrast, the Cohesion Ratio aggregates intra-cluster structure at the global level. Its intra-cluster term \( \mu_I \) is defined as the average similarity across all unordered pairs of points within the same cluster. Unlike \( a(i) \), \( \mu_I \) is not computed per point and is unaffected by individual point geometry. It offers a \emph{shape-agnostic, distribution-independent} summary of global cluster cohesion. This distinction is especially important for clustering in \emph{text embeddings} derived from transformer-based models where semantic similarity does not always manifest as compact geometric clusters. In such spaces, the Silhouette Score may penalize well-formed clusters that are non-convex or overlapping, while the Cohesion Ratio remains robust to such structures, focusing solely on \emph{internal semantic tightness}. This makes \( \rho_C \) a more interpretable for evaluating clustering quality in high-dimensional, meaning-centered domains like natural language.

The Cohesion Ratio can also be interpreted from a null model perspective \cite{hubert1985comparing}. The global similarity \( \mu_G \) serves as the expected intra-cluster similarity under a null clustering model, where cluster assignments are independent of the data distribution. Thus, a ratio \( \rho_C > 1 \) indicates that the clustering exhibits stronger internal cohesion than would be expected under random assignments, suggesting meaningful structure beyond chance.

\subsubsection{Computational Complexity Analysis}

\paragraph{Cohesion Ratio.}
The computation of the Cohesion Ratio \( \rho_C \) involves two main components: the global similarity \( \mu_G \) and the intra-cluster similarity \( \mu_I \). 

The global similarity is defined as
\[
\mu_G = \frac{1}{\binom{n}{2}} \sum_{1 \le i < j \le n} S_{ij}.
\]

which requires evaluating the similarity for all \( \binom{n}{2} \in \mathcal{O}(n^2) \) unordered pairs. In practice, these similarities can be stored in a precomputed similarity matrix.

The intra-cluster similarity \( \mu_I \) is given by
\[
\mu_I = \frac{1}{P} \sum_{k = 1}^K \sum_{\substack{i, j \in C_k \\ i < j}} S_{ij},
\]

where \( P = \sum_{k=1}^K \binom{|C_k|}{2} + N_1 \). In the worst case (e.g., a single cluster), computing \( \mu_I \) also requires \( \mathcal{O}(n^2) \) operations. However, for typical clusterings with multiple clusters, the number of intra-cluster pairs is substantially smaller, yielding faster runtimes in practice. The overall complexity for computing \( \rho_C \) is therefore \( \mathcal{O}(n^2) \), though this can be reduced in practice via memoization and parallelism over cluster-level subcomputations.

\paragraph{Silhouette Score.}
The worst-case complexity of the Silhouette Score is also \( \mathcal{O}(n^2) \) \cite{rousseeuw1987silhouettes}. However, the pointwise formulation limits opportunities for vectorization or efficient reuse of intermediate computations.

\paragraph{Comparison.}
While both metrics have worst-case complexity of \( \mathcal{O}(n^2) \), they differ in computational profile:

\begin{itemize}
    \item The \textbf{Cohesion Ratio} is computed via a global aggregation over a fixed set of pairs, allowing use of precomputed similarity matrices and efficient batching or vectorization.
    \item The \textbf{Silhouette Score} requires pointwise access to intra- and inter-cluster distances, making it more expensive to compute and less amenable to parallelization.
\end{itemize}

Consequently, the Cohesion Ratio is typically faster and more scalable in practice, particularly for large datasets with many small clusters or sparse similarity structures.

\newpage
\section{Metric Profiles Across Cluster Counts}
\label{apx:profiles}

\begin{figure*}[h]
    \centering
    \includegraphics[width=0.8\linewidth]{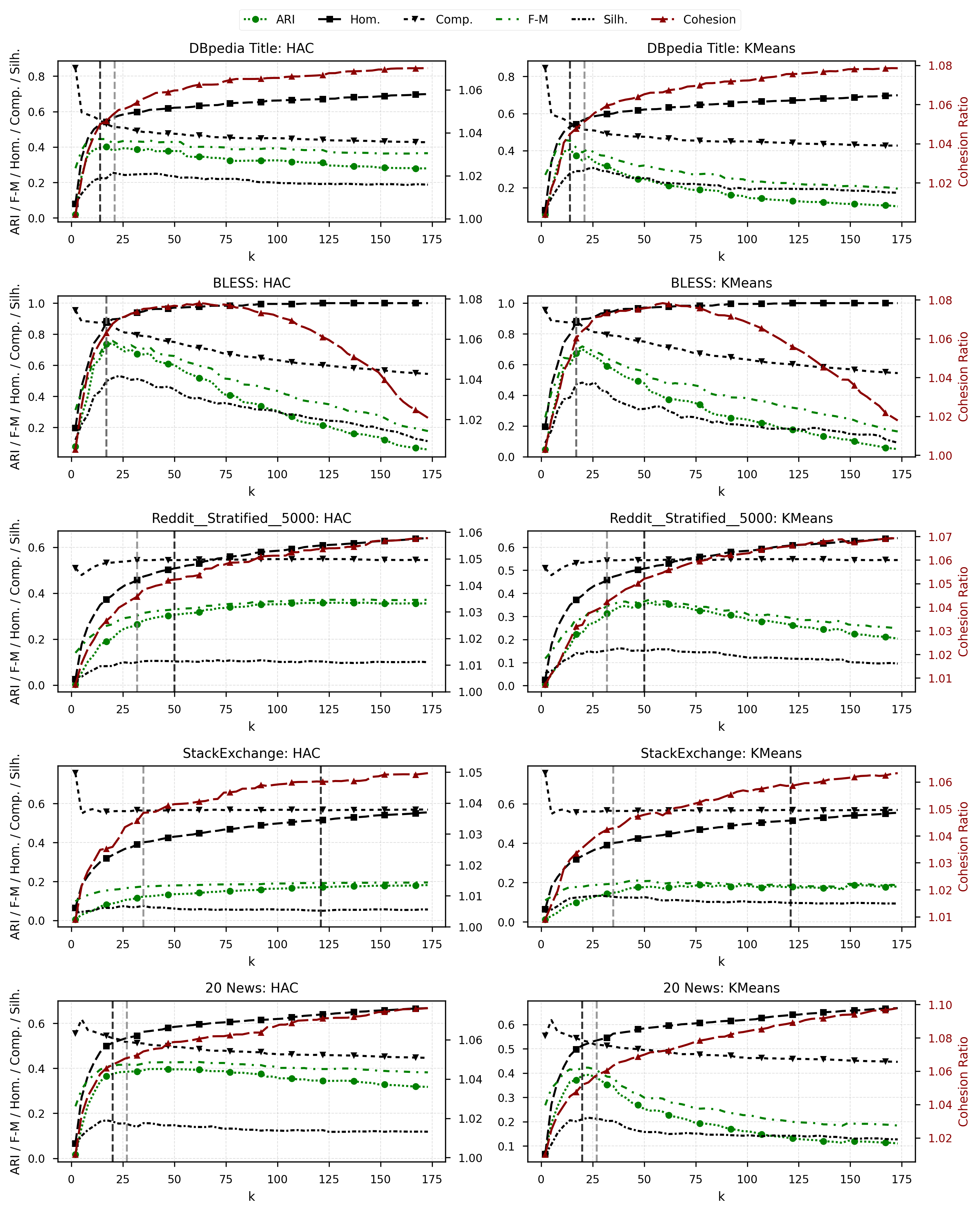}
\caption{
        Variation of clustering metrics (ARI, Homogeneity, Completeness, Fowlkes-Mallows, Silhouette, and Cohesion Ratio) across different numbers of clusters for HAC and K-Means for selected datasets (stratified samples of approx. 5000 data points for each dataset and full BLESS dataset). Each row corresponds to a dataset; each column corresponds to a clustering method. The vertical dashed lines indicate the true (black) and predicted (gray) number of clusters. Metrics are plotted to analyze alignment with ground truth and to highlight the relationship between extrinsic and intrinsic evaluations.
    }
    \label{fig:zscore-heatmaps}
\end{figure*}

\end{document}